\documentclass[11pt,a4paper]{article}
\usepackage[T1]{fontenc}
\usepackage[utf8]{inputenc}
\usepackage{lmodern}
\usepackage[margin=1in]{geometry}
\usepackage{microtype}
\usepackage{amsmath,amssymb,amsthm}
\usepackage{graphicx}
\graphicspath{{fig/}}
\usepackage{booktabs,tabularx,array}
\usepackage[numbers,sort&compress]{natbib}
\usepackage[hidelinks]{hyperref}
\hypersetup{pdftitle={TraceRelay: Attention-Aligned Recurrence over Rolling Traces},
  pdfsubject={Controlled recurrent sequence-model experiments}}
\usepackage{longtable}

\newcommand{\model}{TraceRelay}
\newcommand{\wrap}{\operatorname{wrap}}
\newcommand{\attn}{\operatorname{Attn}}

\newcommand{\paperfigure}[1]{\includegraphics[width=\linewidth]{#1}}
\title{TraceRelay: Attention-Aligned Recurrence over Rolling Traces}
\author{%
  Sungwoo Goo$^{1}$ \qquad Hwi-yeol Yun$^{1}$ \qquad Sangkeun Jung$^{2}$\\
  \small $^{1}$College of Pharmacy, Chungnam National University\\
  \small $^{2}$Department of Computer Science \& Engineering, Chungnam National University\\
  \small Daejeon, Republic of Korea\\
  \small \texttt{swgoo91@gmail.com, hyyun@cnu.ac.kr, hugmanskj@gmail.com}%
}
\date{}

\begin{document}
\maketitle
\begin{abstract}
We present \model, an attention-aligned recurrent architecture that distributes persistent representations over a rolling sequence of low-dimensional traces.
Local right-looking attention forms increments from lower-layer representations; delivery is delayed until all attended inputs are in the causal past.
A fixed additive phase recurrence accumulates the delayed increments, and left-looking attention reads the resulting residual-augmented stream.
A stride-wise prefix sum supports parallel prefill and bounded-buffer continuation.
We study 36 small-model runs on Equal Repeats, bounded Dyck closing-type prediction, and causal Most-Freq generation, using three seeds per setting.
At trained length 256, Equal Repeats models with recurrent phase inheritance reach 98.81--99.69\% accuracy versus 50.73--51.63\% for separately trained variants without inheritance, despite the latter receiving more updates.
Accuracy drops sharply at lengths beyond the training range.
At the longest evaluated lengths, models with more dimensions in the middle layer's recurrent traces perform better on Dyck (76.34\% versus 55.49\% close accuracy at length 4096), whereas models with fewer trace dimensions perform better on five-symbol Most-Freq (70.74\% versus 55.60\% exact generation at length 1024).
These contrasting cases motivate further study of how the size of recurrent representations should be chosen for different tasks, without establishing a general rule across tasks or model configurations.

\begingroup
\renewcommand{\thefootnote}{}
\footnote[0]{Code and model checkpoints are available at
\url{https://github.com/swgoo/trace_relay}.}
\endgroup
\end{abstract}

\section{Introduction}
\label{sec:introduction}
Attention provides a direct interface to multiple past representations, whereas recurrence transports information through time.
The Transformer demonstrated that attention-based sequence computation can be trained without a token-wise recurrent cell \citep{vaswani2017attention}.
However, full attention incurs quadratic attention arithmetic and a key--value cache that grows with the processed context.
Moreover, availability is not utilization: positional sensitivity in \emph{Lost in the Middle} and the length-dependent failures studied by RULER show that accepting a longer input does not guarantee effective use of it \citep{liu2023lost,hsieh2024ruler}.
These findings do not imply that attention bandwidth is irrelevant; they motivate separating accessible history from useful persistent information.

Recurrent and state-space models provide another allocation of that information.
Mamba and Mamba-2 demonstrate input-dependent state updates and efficient parallel formulations of structured recurrence \citep{gu2023mamba,dao2024ssm}.
Nevertheless, recurrence and attention need not cooperate automatically.
SWAX reports that larger sliding windows can weaken long-term memory learning, while excessively small windows harm tasks benefiting from local context \citep{cabannes2025swax}.
Thus, removing access and providing unrestricted access can each create difficulties, for different reasons.

We ask a narrower architectural question: \emph{can recurrence be organized around attention, so that attention learns what to write and retrieve while the temporal transition remains simple?}
\model\ answers with a rolling trace organization.
It forms a candidate update using a short right context, delivers the update only after a causal delay, accumulates it as a phase increment, and exposes the resulting representation to ordinary left-window attention.
Memory formation is available throughout the input stream, rather than being activated only when a downstream query requests recall.

The accompanying design hypothesis is \emph{long in temporal reach, narrow in representational bandwidth}.
A longer read window gives task loss access to intermediate recurrent representations; a narrower projected channel limits the capacity of that same forward path.
H-Net offers complementary motivation for non-uniform capacity allocation: it compresses sequence resolution and allocates richer computation to inner stages \citep{hwang2025hnet}.
Our proposed hierarchy instead preserves the temporal grid while varying trace width and read horizon.

This study contributes the explicit delayed-write/phase-accumulation/left-read graph, its causal and parallel execution account, and a controlled evaluation protocol.
The empirical questions are whether phase inheritance contributes beyond the corresponding ablation without inheritance, whether bounded structured state is usable, and whether a compact aggregate can support autoregressive generation.

\section{Attention-aligned rolling recurrence}
\label{sec:method}
\subsection{Notation}
Let $h_t^{\ell-1}\in\mathbb{R}^{H}$ be the input to layer $\ell$ at sequence position $t$, counted from zero at the start of the sequence.
Within a layer, write $h_t=h_t^{\ell-1}$; $h_t^\ell$ denotes its output.
The trace width is $d$, the relay stride is $W$, and $L$ and $R$ count the preceding and following positions visible to Left and Right attention, respectively.
Both windows include their center, so their maximum sizes are $L+1$ and $R+1$; causality requires $W>R$.

For readability, RMSNorm is implicit in the equations: it is applied to the lower-layer stream before Right attention, the memory-augmented stream before Left attention, and the post-attention state before the FFN.
Residual additions use the corresponding unnormalized states.
We use $\attn_R(\cdot,\cdot)$ and $\attn_L(\cdot,\cdot)$ for complete local multi-head attention outputs: the first argument supplies queries and the second supplies keys and values.
Both outputs have width $d$.

The learned bias-free maps are $P_\delta:\mathbb{R}^{d}\to\mathbb{R}^{d}$, $P_\phi:\mathbb{R}^{2d}\to\mathbb{R}^{H}$, and $P_L:\mathbb{R}^{d}\to\mathbb{R}^{H}$; $P_L$ maps the Left output into the residual stream, separately from the attention operator's internal output projection.
The fixed phase-update scale $\alpha>0$ is $\pi$ in the executed experiments, and $\wrap(a)=a\bmod 2\pi$ acts elementwise.
The initial phase is $\theta_{t<0}=0$.
Intermediate variables below denote the Right output $o_i^{\mathrm R}$, bounded increment $\delta_i$, delivered increment $u_t$, phase $\theta_t$, memory-augmented state $m_t$, Left output $o_t^{\mathrm L}$, and post-attention state $\widetilde h_t$.
The upright superscripts on $o_i^{\mathrm R}$ and $o_t^{\mathrm L}$ label attention paths, not layers.
For the complex-coordinate interpretation below, $z_t=\exp(i\theta_t)$.

\subsection{Delayed right-looking writes}
The Right writer forms a candidate from the lower-layer stream:
\begin{equation}
 o_i^{\mathrm R}=\attn_R\bigl(h_i,h_{i:i+R}\bigr)\in\mathbb{R}^{d},\qquad
 \delta_i=\alpha\tanh\bigl(P_\delta(o_i^{\mathrm R})\bigr)\in\mathbb{R}^{d}.
 \label{eq:writer}
\end{equation}
The elementwise $\tanh$ bounds each coordinate of $\delta_i$ in magnitude by $\alpha$.
At arrival position $t$, the delivered increment is
\begin{equation}
 u_t=\begin{cases}\delta_{t-W},&t\geq W,\\0,&t<W.\end{cases}
 \label{eq:delay}
\end{equation}
Right-attention keys and centers use the input validity mask; the relay advances over sequence positions, including padding.
The implementation computes these writes directly at their arrival positions by shifting query centers.
It does not require an unavailable future input during decoding.

\begin{figure}[htbp]
\centering
\paperfigure{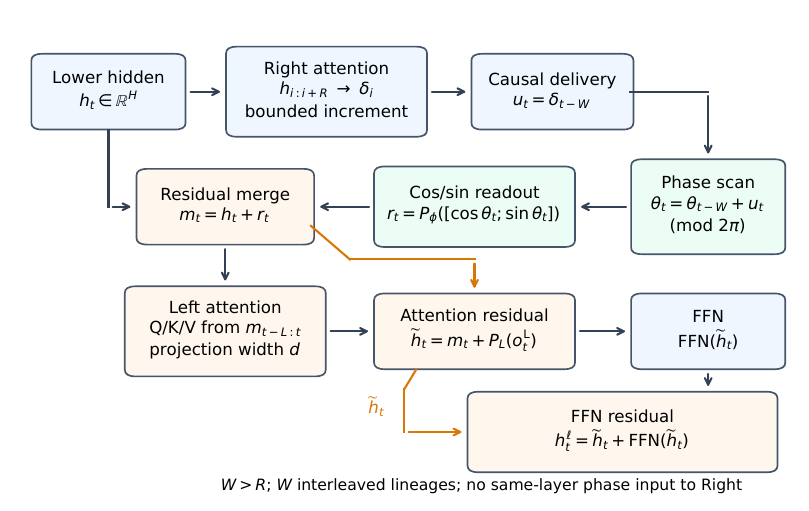}
\caption[Architecture]{\textbf{Reference computation graph.} Right attention observes lower-layer positions $i$ through $i+R$; its increment is delivered at $i+W$, with $W>R$. The stride-wise phase scan and cos/sin readout precede the residual merge and Left attention. The displayed readout applies to initialized lineages; until a valid delayed update arrives, the readout is zero. Left Q/K/V consume the merged stream. The orange branches pass the merged representation $m_t$ to the attention residual and the post-attention state $\widetilde h_t$ to the FFN residual. Right does not read this layer's phase. Normalization is implicit, as in the equations.}
\label{fig:architecture}
\end{figure}

\subsection{Fixed phase accumulation, learned increments}
The recurrent component is
\begin{equation}
 \theta_t=\wrap(\theta_{t-W}+u_t).
 \label{eq:phase_accumulation}
\end{equation}
There are $W$ interleaved trace lineages.
After warm-up, a new update can arrive at \emph{every token};
$W$ is the interval between successive states of one lineage.

In complex coordinates, Eq.~\eqref{eq:phase_accumulation} is coordinate-wise multiplication by $\exp(iu_t)$.
The transport law has no learned decay or dense state-transition matrix.
The increments are nevertheless learned and input-dependent.

\subsection{Left attention reads a recurrently augmented stream}
At positions whose recurrent lineage has received a valid delayed update, the layer computes
\begin{align}
 m_t &= h_t + P_\phi\bigl([\cos\theta_t;\sin\theta_t]\bigr),\label{eq:merge}\\
 o_t^{\mathrm L} &= \attn_L\bigl(m_t,m_{t-L:t}\bigr),\label{eq:reader}\\
 \widetilde h_t &= m_t+P_L(o_t^{\mathrm L}),\qquad
 h_t^{\ell}=\widetilde h_t+\operatorname{FFN}(\widetilde h_t).
 \label{eq:layer}
\end{align}
The executed experiments use zero attention and residual dropout.

Crucially, the left reader attends to the \emph{merge of lower-layer hidden features and phase readout}, not to a phase-only bank.
Trace width controls attention projections and phase width; it is not the full residual-stream width.
Information read in layer $\ell$ can condition a writer in layer $\ell+1$.
This gives cross-layer refinement without a same-layer state-dependent writer.
The rolling collection acts as distributed working state rather than a separate slot store.

\paragraph{Initialization and masked inputs.}
Until a lineage receives its first valid delayed update, its phase readout is set to zero, so $m_t=h_t$.
For an unpadded sequence, this applies to the first $W$ positions.
This condition prevents the nonzero $\cos(0)$ features of an uninitialized phase from entering the residual stream.
The implementation tracks it with Boolean flags: $q_t$ marks a valid delivered source center, and $v_t=v_{t-W}\lor q_t$, initialized with $v_{t<0}=0$, records whether the lineage has received any valid update.
An invalid source center contributes a zero increment; it cannot initialize a lineage or clear an existing validity flag.
A valid update initializes the lineage even when its increment is zero.
The implementation masks the cos/sin features when $v_t=0$; the bias-free $P_\phi$ then produces exactly zero.

\subsection{Causality and parallel execution}
The latest lower-layer position used by $u_t$ is $t-W+R<t$.
Therefore $m_t$ depends only on inputs through $t$, and the left reader accesses only $m_j$ for $j\leq t$.
Induction over layers gives a causal decoder; logits at $t$ may predict token $t+1$.

For each residue $r\in\{0,\ldots,W-1\}$,
\begin{equation}
 \theta_{r+nW}=\wrap\left(\sum_{k=0}^{n}u_{r+kW}\right).
 \label{eq:scan}
\end{equation}
The complete lower-layer stream determines all increments before the scan.
Thus a reshape, cumulative sum, and phase wrap implement the recurrence during prefill.
Continuation adds the saved phase tail to each lineage.
The core uses standard tensor operations and local attention, with an optional FlashAttention-2 backend \citep{dao2023flash2}.

\subsection{Why long and narrow?}
\label{sec:longnarrow}
A rolling reader can send task gradients directly to several intermediate trace-bearing representations instead of relying only on the latest recurrent state.
Our hypothesis is that this access makes useful writes easier to learn.
Narrowing the projected channel may simultaneously limit an attention-only solution that bypasses persistence.
Both effects depend on the task and training distribution:
a backward path is also a forward path, and even a narrow channel can encode all information needed for a compact task.

We therefore treat $L$, $R$, $W$, and $d$ as distinct controls.
A short $R$ permits a short causal delay and repeated within-lineage updates during training; increasing $R$ forces the lower bound on $W$ upward.
An inner long-narrow path can expose more trace generations without proportionally increasing projected width. 
This is a design hypothesis, not a proof that narrowing is necessary or that widening prevents learning.
In particular, the reference experiments below vary center width with fixed Left attention windows; they do not independently establish the benefit of a longer center window.

\section{Controlled evaluation}
\label{sec:experiments}
All three tasks use one-hot symbols, a bias-free input projection, and the same reference core.
The executed sweeps use $H=64$, FFN width $256$, three layers, four heads, and trace widths $[64,d_c,64]$ with $d_c\in\{16,32,64\}$.
Each layer's Left attention sees up to 15 preceding positions, Right attention sees up to 7 following positions, and the relay stride is 8; both windows include the center.
Training seeds are $42,43,44$ for each setting.
The width comparisons characterize task-specific behavior, with parameter count and optimization exposure also varying across widths.

\paragraph{Equal Repeats: recurrent inheritance.}
We follow the historical Equal Repeats semantics in the code accompanying \citet{deletang2023chomsky}.
Inputs have the form $0^a1^b0^c$ with a guaranteed equality between run lengths; the target identifies the corresponding relation.
Inclusive sampling preserves cases with an empty run and assigns the all-equal case to class~0.
Training samples a uniform integer length from $32$ through $256$.
A classifier reads the final valid position.
Models with and without phase inheritance are separately initialized and trained for every width and seed.
The ablation without inheritance retains the writer, delay, phase features, and reader, replacing Eq.~\eqref{eq:phase_accumulation} with $\theta_t=\wrap(u_t)$ and $v_t=q_t$ throughout training and evaluation.
Paired protocols match in task, geometry, parameter count, training seed, and final-example fingerprints; their backbone configurations differ only in whether the previous phase is inherited.

\paragraph{Bounded Dyck: structured prediction.}
Dyck-$(k,m)$ contains $k$ bracket types and nesting depth at most $m$ \citep{hewitt2020dyck}.
The executed phase-inheritance sweep uses $k=8,m=10$, with a uniform even training length from $32$ through $256$.
Completion-count dynamic programming samples bounded paths of the chosen length, and opening types are sampled uniformly.
Logits at $t$ predict the closing type at $t+1$; opening targets and padding are excluded.
No stack metadata is supplied to the model.
We report closing-type token accuracy by input length, matching-opener distance, and depth, with counts for each slice.
This diagnostic measures structured prediction within the bounded language rather than full-language likelihood or unbounded-stack recognition.

\paragraph{Most-Freq: aggregate-conditioned generation.}
We adapt Most-Freq from \citet{weiss2021rasp}: emit present symbols in descending frequency, breaking ties by first appearance.
The phase-inheritance sweep uses five source symbols and uniform integer source lengths from $128$ through $256$.
The separated-count sampler makes every count positive, enforces a minimum adjacent count gap of four, and randomizes symbol assignment and token order; it is not uniform over count combinations.
Consequently, this sweep has five-symbol rankings and does not exercise the tie rule.
Training uses shifted teacher-forced targets.
Final evaluation starts from source plus separator and greedily feeds back only the model's own symbols.
Exact correctness requires the complete ranking followed by EOS, for six output tokens.
Ranking, set-validity, and termination diagnostics and shortcut baselines use the same generated examples. This is a short-output autoregressive transduction task.

\paragraph{Selection and reporting.}
Checkpoint selection uses the unweighted mean across ID validation lengths: classification accuracy for Equal Repeats, closing-type accuracy for Dyck, and source-only free-generation exact accuracy for Most-Freq.
Ties use lower mean validation cross-entropy (teacher-forced cross-entropy for Most-Freq).
All runs have a $20{,}000$-update cap and stop after two consecutive validation means reach $99\%$, with checks every $250$ updates.
OOD results do not control checkpoint selection or stopping.
The repeated threshold uses the same validation examples and is a stopping rule, not an independent replication.

All $36$ planned final runs are retained: nine Equal Repeats runs with phase inheritance and nine without, nine Dyck runs, and nine Most-Freq runs.
Every run completed either by the threshold or by its update cap.
We report individual training seeds and their unweighted mean with sample standard deviation.
Final-example fingerprints match across variants and seeds within each task, so the same examples are reused when comparing models.
Training exposure is unequal: the recorded optimizer steps, tokens, selected-checkpoint steps, and wall times are reported in Table~\ref{tab:exposure}; detailed executed protocols appear in Table~\ref{tab:protocol} and Appendix~\ref{app:protocol}.

\section{Results}
\label{sec:results}
All results below use the checkpoint selected by in-distribution (ID) validation.
Evaluation inputs are shared across training seeds and widths within a task.
Error bars are sample standard deviations across three training seeds, not confidence intervals over independent evaluation datasets.
Table~\ref{tab:main_results} uses common comparison lengths 256 and 1024; the figures retain every tested length.

\begin{figure}[htbp]
\centering
\paperfigure{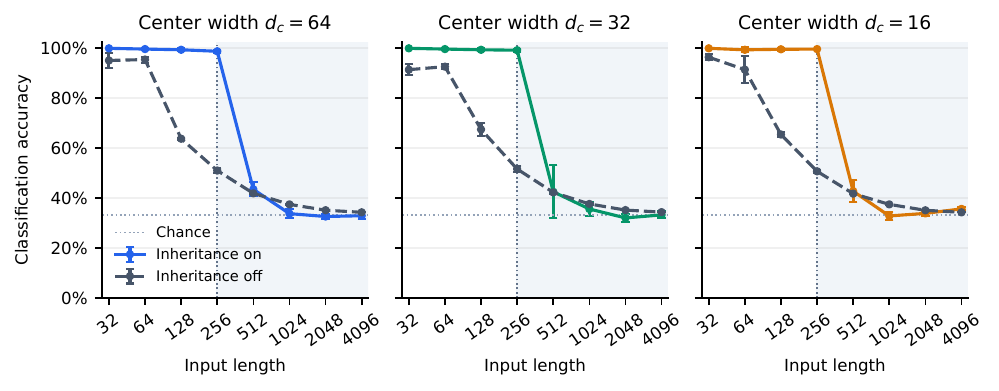}
\caption[Equal Repeats]{\textbf{Equal Repeats: with and without phase inheritance.} Accuracy on 10,000 common examples per length, mean $\pm$ sample SD over three training seeds. Shading marks lengths beyond the 32--256 training range. Models without inheritance are separately trained with the same task and geometry; all use 20,000 updates, while models with inheritance stop earlier. The strong separation within the trained range largely disappears in OOD evaluation. The dotted line is the $1/3$ chance reference.}
\label{fig:equal_repeats}
\end{figure}

\paragraph{Does recurrent inheritance contribute?}
At $T=256$, models with phase inheritance reach 98.81, 99.24, and 99.69\% for center widths 64, 32, and 16, respectively; matched models without inheritance reach 51.03, 51.63, and 50.73\% (Figure~\ref{fig:equal_repeats}).
The corresponding mean paired gains are 47.78, 47.61, and 48.97 percentage points.
The graph without inheritance retains the delayed writer and reader but has maximum single-pass input lag 69, compared with unrestricted recurrent ancestry when the phase is inherited.
All runs without inheritance consume their 20,000-update budget, whereas those with inheritance stop after 4,250--10,250 updates.
The result supports the utility of inheritance on this training distribution despite greater optimization exposure for the ablation.

That benefit does not imply a learned length-invariant counting rule.
At $T=512$, the models with inheritance average only 42.52--43.51\%, and the paired mean gains shrink to 0.24--1.61 points.
At $T=1024$--$4096$, both variants are close to the three-class chance reference; the ablation without inheritance even has the higher mean at 1024 for all widths.
We retain this failure alongside the strong ID separation.

\paragraph{How does width affect structured prediction?}
Bounded Dyck exhibits a different pattern.
At $T=1024$, mean close-type accuracies are 93.03, 89.48, and 87.17\% for widths 64, 32, and 16.
At $T=4096$, these decline to $76.34\pm1.81$, $63.47\pm8.36$, and $55.49\pm10.49$\%, respectively.
The wider center therefore has the highest mean length-OOD accuracy in this sweep.
Eight runs meet the stopping rule; width 16/seed 43 reaches the full budget and remains included.

\begin{figure}[htbp]
\centering
\paperfigure{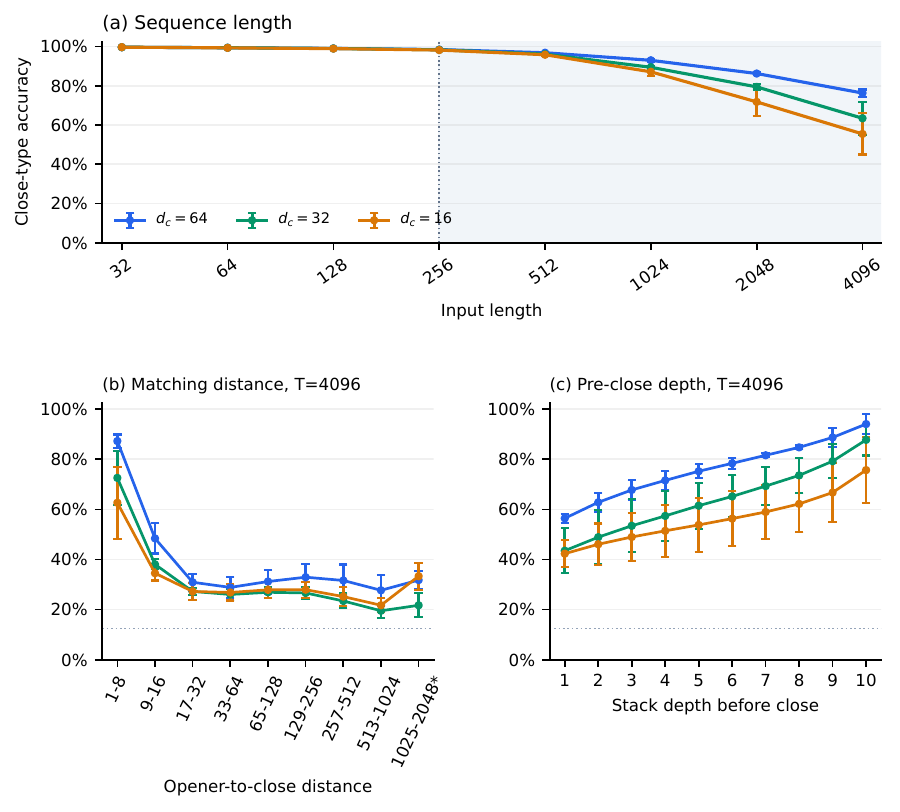}
\caption[Dyck]{\textbf{Bounded Dyck closing-type prediction.} Three-seed mean $\pm$ sample SD for $k=8,m=10$, using 2,000 common sequences per length. (a) Close accuracy versus length; shaded lengths exceed the training maximum 256. (b,c) Matching-distance and pre-close-depth slices at $T=4096$, with 4,096,000 close targets in total. The starred distance bin 1025--2048 has only 43 targets; the empty $>2048$ bin is omitted. The 513--1024 bin has 1,639 targets. Dotted horizontal lines in (b,c) mark 12.5\% chance. Full slice counts are in Table~\ref{tab:dyck_slices}.}
\label{fig:dyck}
\end{figure}

Overall accuracy is dominated by short dependencies: 77.98\% of evaluated closes at $T=4096$ have matching distance 1--8.
At distance 513--1024, width 64 achieves $27.70\pm6.14$\%, above the 12.5\% type-chance reference, on 1,639 closes.
The next bin contains only 43 closes, and no example has distance above 2048.
Figure~\ref{fig:dyck} and Table~\ref{tab:dyck_slices} distinguish sequence length, opener distance, and pre-close depth.
The distance-slice ordering is not uniformly monotone in width; sequence-length accuracy is not a measurement of a 4096-token memory lifetime.

\paragraph{Can the state support ordered generation?}
Most-Freq reverses the mean width ordering observed on Dyck.
At $T=512$, exact ranking-plus-EOS accuracy is $87.76\pm3.73$, $93.55\pm1.53$, and $94.79\pm0.59$\% for widths 64, 32, and 16.
At $T=1024$, the corresponding scores are $55.60\pm12.82$, $65.40\pm2.39$, and $70.74\pm4.95$\%.
Width 64 has particularly large seed variation: 43.65--69.14\% at 1024.

At $T=1024$, Top-1 accuracy remains 86.82--90.76\%, higher than full-output exact accuracy.
Termination accuracy ranges from 82.91\% for width 64 to 96.22\% for width 32.
Thus ranking and output-protocol errors both matter (Table~\ref{tab:most_freq_diagnostics}).
On the same inputs, the 32-token suffix-frequency baseline reaches 1.27\% full exact, first appearance 1.07\%, and fixed symbol order 0.59\%.
These reject those specific shortcuts, without substituting for a separately trained comparison without phase inheritance.
\begin{figure}[htbp]
\centering
\paperfigure{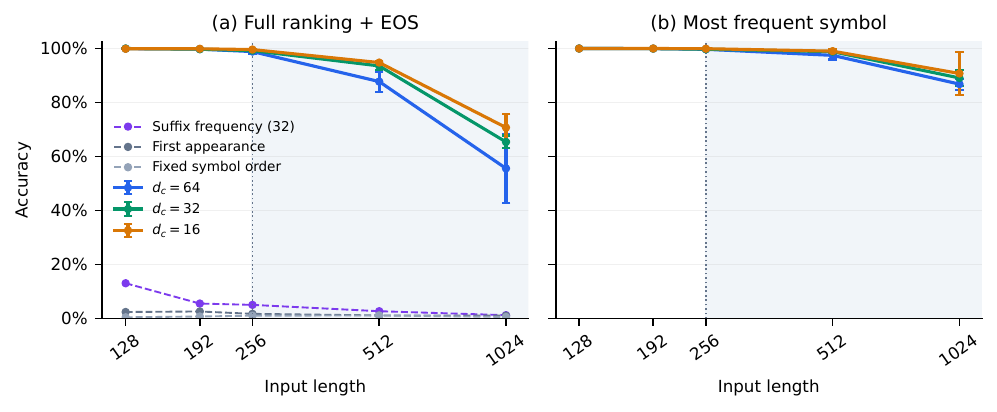}
\caption[Most-Freq]{\textbf{Most-Freq causal generation.}
Five symbols, separated counts with minimum gap four, training lengths 128--256, and 1,024 common evaluation examples per length.
Curves show mean $\pm$ sample SD over three training seeds; shading denotes length OOD.
(a) Exact greedy ranking including predicted EOS, alongside same-input shortcut baselines. (b) Top-1 accuracy.
\emph{Suffix frequency (32)} ranks symbols present in the last 32 input tokens by descending frequency there, breaking ties by first occurrence within that suffix.
\emph{First appearance} lists symbols seen in the full input in order of first occurrence, ignoring their counts.
\emph{Fixed symbol order} always lists $0,1,2,3,4$.
All three heuristics append EOS and use the same evaluation inputs.
No gold output is fed back during model generation.
Narrower centers have higher OOD means in this sweep, with unequal early-stopped training exposure.}
\label{fig:most_freq}
\end{figure}

\paragraph{Optimization and summary.}
Figure~\ref{fig:optimization} shows all unsmoothed ID validation histories, with selected checkpoints marked.
Most-Freq width 64 stops after a mean 2,750 updates, compared with 4,000 for width 32 and 3,667 for width 16.
Faster attainment of the ID threshold therefore does not coincide with the best OOD score.
Width also changes parameter count, and early stopping changes token exposure; Table~\ref{tab:exposure} records both.
These are matched-protocol comparisons, not equal-compute experiments.
\begin{table}[htbp]
\centering\small
\caption{Best-checkpoint test results, mean $\pm$ sample SD over seeds 42--44 (percent). ID is $T=256$ and OOD is $T=1024$ for every task. Metrics are sequence classification (Equal Repeats), token-weighted close type (Dyck), and exact generated ranking including EOS (Most-Freq).}
\label{tab:main_results}
\begin{tabular}{llccc}
\toprule
Task & Phase inheritance & $d_c$ & ID: 256 & OOD: 1024 \\
\midrule
Equal Repeats & On & 64 & $98.81\pm0.58$ & $33.76\pm1.82$ \\
Equal Repeats & On & 32 & $99.24\pm0.56$ & $35.63\pm2.99$ \\
Equal Repeats & On & 16 & $99.69\pm0.13$ & $32.84\pm1.66$ \\
Equal Repeats & Off & 64 & $51.03\pm0.94$ & $37.48\pm0.14$ \\
Equal Repeats & Off & 32 & $51.63\pm1.24$ & $37.65\pm0.23$ \\
Equal Repeats & Off & 16 & $50.73\pm0.52$ & $37.47\pm0.09$ \\
\midrule
Dyck & On & 64 & $98.52\pm0.05$ & $93.03\pm0.69$ \\
Dyck & On & 32 & $98.40\pm0.12$ & $89.48\pm0.82$ \\
Dyck & On & 16 & $98.26\pm0.17$ & $87.17\pm2.31$ \\
\midrule
Most-Freq & On & 64 & $98.86\pm0.78$ & $55.60\pm12.82$ \\
Most-Freq & On & 32 & $99.22\pm0.35$ & $65.40\pm2.39$ \\
Most-Freq & On & 16 & $99.61\pm0.20$ & $70.74\pm4.95$ \\
\bottomrule
\end{tabular}
\end{table}

\clearpage
\section{Discussion and scope}
\label{sec:discussion}
\paragraph{An organization of recurrence, not a new primitive.}
The distinctive design choice is the combination of right-context-conditioned delayed writes, same-layer-independent increment generation, interleaved phase transport, and a left reader over the recurrently augmented rolling stream.
Attention and recurrence individually, and their combination, are not new.
Block-Recurrent Transformers use attention and gates over a set of block-level states, while Recurrent Memory Transformer passes memory tokens between segments \citep{hutchins2022block,bulatov2022rmt}.
TransformerFAM reuses latent representations through feedback \citep{hwang2024fam}; Maglev combines sliding recurrent memory with prefiller-generated targets and a consistency objective \citep{liu2026maglev}.
Our contribution should be compared at the dependency-graph level, not described as the first use of attention over recurrent states.

Mamba-2 motivates designing recurrence around an efficient algebraic execution form \citep{dao2024ssm}.
Gated DeltaNet also combines structured updates with a parallel algorithm \citep{yang2024gdn}.
Mamba-3 provides independent evidence for complex-valued state updates in state tracking \citep{lahoti2026mamba3}; it does not validate \model's particular phase update or imply that all such recurrences train equally well.
Our fixed transport law leaves learned selection and content formation in attention and projections.

\paragraph{Bandwidth and reach are different controls.}
The intended long-narrow center is a proposed compromise between exposing intermediate states and limiting direct-access capacity.
SWAX motivates the interaction between attention access and memory learning; H-Net motivates non-uniform representation allocation \citep{cabannes2025swax,hwang2025hnet}.
Our executed sweep changes width while every Left reader still sees at most 15 preceding positions.
At the longest evaluated lengths, mean OOD performance favors wider centers on Dyck and narrower centers on Most-Freq.
This contrast offers a useful case study of task-specific bandwidth preferences in compact frequency aggregation and bounded nested-state prediction, without establishing a general rule across tasks or configurations.

\paragraph{What remains untested.}
This study does not measure language-model quality, agent performance, or compute-matched long-context scaling.
Familiar primitives reduce implementation novelty but do not guarantee quality or hardware efficiency in a new arrangement.
Width changes also affect parameter count and optimization, and early-stopped runs can have different training exposures. Bounded phase values do not guarantee reliable retention at arbitrary horizons.
\section{Conclusion}
\model\ reorganizes recurrence around attention: a right-looking writer prepares causally delayed increments, a simple phase scan transports them, and a left-looking reader accesses their rolling representations.
This yields a token-grid-preserving architecture with parallel prefill and bounded inference buffers, without an explicit slot-commit subsystem.
Equal Repeats demonstrates a strong contribution from recurrent inheritance inside the training range but fails to establish reliable length extrapolation.
At the longest evaluated lengths, mean OOD performance favors wider centers on Dyck and narrower centers on Most-Freq.
These contrasting cases motivate further study of task-specific allocation of representational bandwidth.
The controlled results do not identify a universal optimal width or an efficiency advantage. Separately varying read horizon, writer width, and training exposure is a concrete next step toward testing the long-narrow design hypothesis.

\clearpage
\bibliographystyle{unsrtnat}
\bibliography{references}
\clearpage
\appendix
\section{Executed task protocols and diagnostics}
\label{app:protocol}
\begin{table}[htbp]
\centering\small
\setlength{\tabcolsep}{4pt}
\renewcommand{\arraystretch}{1.12}
\caption{Executed protocols. All tasks sweep $d_c\in\{64,32,16\}$ with three training seeds per variant and width. ID validation occurs every 250 updates; two consecutive mean ID scores $\geq99\%$ trigger stopping. OOD results never control training. Table~\ref{tab:exposure} gives individual stopping/selected steps and token exposure.}
\label{tab:protocol}
\begin{tabularx}{\linewidth}{@{}>{\raggedright\arraybackslash}p{.20\linewidth}*{3}{>{\raggedright\arraybackslash}X}@{}}
\toprule
Field & Equal Repeats & Dyck & Most-Freq \\
\midrule
Phase inheritance & On and off; 18 runs & On; 9 runs & On; 9 runs \\
Task parameters & Binary $0^a1^b0^c$; three relation classes & $k=8$ bracket types; maximum depth $m=10$ & $V=5$; decreasing count, first-appearance ties \\
Sampler & Inclusive repeated length $r\in[1,\lfloor T/2\rfloor]$; equal-all maps to class 0 & Completion-count DP over bounded paths; uniform opening types & Separated positive counts, gap $\geq4$; randomized symbol assignment and order \\
Online source length & Uniform integer $T\in[32,256]$ & Uniform even $T\in[32,256]$ & Uniform integer $T\in[128,256]$ \\
ID validation lengths & 32, 64, 128, 256 & 32, 64, 128, 256 & 128, 192, 256 \\
Final ID lengths & 32, 64, 128, 256 & 32, 64, 128, 256 & 128, 192, 256 \\
Final OOD lengths & 512, 1024, 2048, 4096 & 512, 1024, 2048, 4096 & 512, 1024 \\
Examples per length (validation / final) & 1,024 / 10,000 & 1,024 / 2,000 & 512 / 1,024 \\
Training seeds & 42, 43, 44 & 42, 43, 44 & 42, 43, 44 \\
Validation / final seed & 20261007 / 20261006 & 20261007 / 20261006 & 20261007 / 20261006 \\
Batch size (train / eval) & 64 / 128 & 64 / 64 & 64 / 64 \\
Best-checkpoint score & Mean ID classification accuracy & Mean ID close-type accuracy & Mean ID greedy exact accuracy including EOS \\
Selection tie-break & Lower mean ID CE & Lower mean ID close CE & Lower mean ID teacher-forced CE \\
Update cap / executed & 20,000 / 4,250--20,000 & 20,000 / 14,250--20,000 & 20,000 / 2,500--5,000 \\
Backend / precision & FlashAttention-2 / BF16 & FlashAttention-2 / BF16 & FlashAttention-2 / BF16 \\
Output and EOS & Final-position class; no EOS target & Next closing type only; no EOS target & Source-only greedy generation; five symbols then EOS required \\
\bottomrule
\end{tabularx}
\par\smallskip
\begin{minipage}{\linewidth}\footnotesize
All runs use AdamW ($3\!\times\!10^{-4}$, $\beta=(0.9,0.999)$, $\epsilon=10^{-8}$, weight decay $0.01$), a 100-update linear warmup followed by constant learning rate, and gradient-norm clipping at $1$. Each online batch shares one sampled source length. Evaluation examples match across seeds and variants at each length; reported uncertainty is sample SD across trained seeds.
\end{minipage}
\end{table}

The table records the completed sweeps rather than trainer defaults. All three task adapters use the same fixed core geometry, AdamW, a 100-update linear learning-rate warm-up followed by a constant rate, and batch size 64. Validation is performed every 250 updates. Changing windows in earlier task pilots and moving from four to eight to five Most-Freq symbols were exploratory calibration decisions, not preregistered choices; the present analysis retains the complete final sweep for each declared protocol.

\paragraph{Equal Repeats semantics.}
The sampled families correspond to $a=b$, $b=c$, or $a=c$ in $0^a1^b0^c$. The historical generator uses inclusive run-length sampling; an even-length endpoint can produce an empty run, and equal-all examples are relabeled with class-zero priority. Thus the measured class distribution, not an assumed exact balance, defines a majority-class baseline. The implementation records evaluation fingerprints independent of inference batching. Our phase-inheritance gaps pair the same task distribution, geometry, seed, and evaluation fingerprints. At the tested power-of-two lengths, the final evaluation contains class counts $(3333,3333,3334)$, giving a 33.34\% majority baseline. Both variants are trained from fresh initialization; they have identical parameter counts within each width.

\paragraph{Bounded Dyck slices.}
The sampler begins and ends with an empty stack, never exceeds $m$, and does not require every sequence to reach depth $m$. Its open/close choices use normalized completion counts, and opening types are uniform. The same $k,m$ must be retained for a length-only comparison. Longer examples may change opener-distance and depth distributions despite a fixed maximum stack depth. Table~\ref{tab:dyck_slices} reports the common-set distance and depth counts at $T=4096$; the accompanying JSON retains each slice at every tested length. Empty bins remain missing, not zero-accuracy observations. Do not identify good close-type accuracy with recovery of every complete latent stack.

\begin{table}[htbp]
\centering\scriptsize
\caption{Dyck diagnostics at $T=4096$: percent close accuracy, mean $\pm$ sample SD. Counts are targets in the common 2,000-sequence evaluation set, not independent replications across models. An empty bin is unavailable, not zero accuracy.}\label{tab:dyck_slices}
\begin{tabular}{lrrrr}
\toprule Slice & Closes & $d_c=64$ & $d_c=32$ & $d_c=16$ \\
\midrule
1--8 & 3,194,038 & $87.27\pm2.62$ & $72.57\pm10.69$ & $62.65\pm14.31$ \\
9--16 & 360,329 & $48.41\pm6.04$ & $38.07\pm1.98$ & $34.52\pm2.93$ \\
17--32 & 257,308 & $30.90\pm3.45$ & $27.24\pm1.32$ & $27.23\pm3.41$ \\
33--64 & 159,952 & $28.85\pm4.23$ & $26.04\pm2.23$ & $26.83\pm3.21$ \\
65--128 & 80,458 & $31.20\pm4.76$ & $26.90\pm2.19$ & $27.89\pm3.08$ \\
129--256 & 32,358 & $32.93\pm5.24$ & $26.64\pm2.36$ & $27.94\pm3.24$ \\
257--512 & 9,875 & $31.62\pm6.37$ & $23.44\pm2.98$ & $25.18\pm3.81$ \\
513--1024 & 1,639 & $27.70\pm6.14$ & $19.52\pm2.88$ & $21.72\pm2.82$ \\
1025--2048 & 43 & $31.78\pm3.55$ & $21.71\pm4.84$ & $33.33\pm5.37$ \\
$>2048$ & 0 & -- & -- & -- \\
\midrule
Depth 1 & 96,108 & $56.43\pm1.80$ & $43.50\pm9.01$ & $42.32\pm5.45$ \\
Depth 2 & 254,874 & $62.82\pm3.81$ & $48.90\pm10.70$ & $46.07\pm8.30$ \\
Depth 3 & 436,164 & $67.75\pm3.87$ & $53.45\pm10.59$ & $48.98\pm9.62$ \\
Depth 4 & 592,750 & $71.56\pm3.66$ & $57.39\pm10.00$ & $51.44\pm10.37$ \\
Depth 5 & 682,795 & $75.21\pm2.83$ & $61.47\pm9.17$ & $53.80\pm10.63$ \\
Depth 6 & 680,971 & $78.38\pm2.15$ & $65.17\pm8.43$ & $56.31\pm10.91$ \\
Depth 7 & 587,399 & $81.59\pm0.82$ & $69.29\pm7.70$ & $58.97\pm10.83$ \\
Depth 8 & 428,589 & $84.75\pm0.74$ & $73.57\pm7.03$ & $62.18\pm11.35$ \\
Depth 9 & 246,573 & $88.70\pm3.75$ & $79.22\pm6.82$ & $66.71\pm11.84$ \\
Depth 10 & 89,777 & $94.06\pm3.90$ & $87.75\pm6.23$ & $75.68\pm13.07$ \\
\bottomrule
\end{tabular}
\end{table}

\paragraph{Most-Freq controls.}
For $V$ symbols and count gap $g$, separated sampling requires $T\geq V+gV(V-1)/2$. It adds sorted random extras to the increasing minimum counts and then independently shuffles symbol labels and positions. This is not uniform sampling over count vectors. Changing $T$ can change realized margins, so length OOD is not an isolated retention test. IID-uniform evaluation, if used, must be labeled as distribution shift. The output ranking has at most $V$ symbols, and EOS is predicted rather than forcibly appended by the model.

The fixed-order, first-appearance, and suffix-frequency controls read different information. A suffix covering the whole source is an oracle case. Exceeding these controls rejects those particular strategies, not every local computation. Greedy free-generation exact accuracy requires the complete ranking and termination; teacher-forced token accuracy does not. For deterministic causal execution, per-example \emph{full-sequence} teacher-forced exact and greedy exact should agree when logits have identical numerical decisions. The BF16 evaluation records 18 TF/greedy exact disagreements across 46,080 model--example evaluations, with at most three in any 1,024-example cell. We preserve these diagnostics and report actual greedy outcomes; we do not infer their numerical cause or adjust scores. Table~\ref{tab:most_freq_diagnostics} separates ranking, set, and termination behavior.

\begin{table}[htbp]
\centering\small
\caption{Most-Freq diagnostics at $T=1024$ (percent; three-seed mean $\pm$ SD). Top-2 requires the first two outputs in the correct order. Set exact requires each source symbol exactly once but ignores rank; termination records a predicted EOS within the decode limit.}
\label{tab:most_freq_diagnostics}
\begin{tabular}{lccc}
\toprule
Metric & $d_c=64$ & $d_c=32$ & $d_c=16$ \\
\midrule
Full exact & $55.60\pm12.82$ & $65.40\pm2.39$ & $70.74\pm4.95$ \\
Top-1 & $86.82\pm2.05$ & $89.10\pm3.00$ & $90.76\pm7.91$ \\
Top-2 prefix & $74.41\pm7.43$ & $78.71\pm2.03$ & $84.05\pm6.15$ \\
Set exact & $81.84\pm7.26$ & $87.96\pm3.97$ & $89.55\pm4.67$ \\
Termination & $82.91\pm7.32$ & $96.22\pm5.36$ & $95.64\pm5.61$ \\
\bottomrule
\end{tabular}
\end{table}

\clearpage
\paragraph{Training histories and uncertainty.}
\begin{figure}[htbp]
\centering
\paperfigure{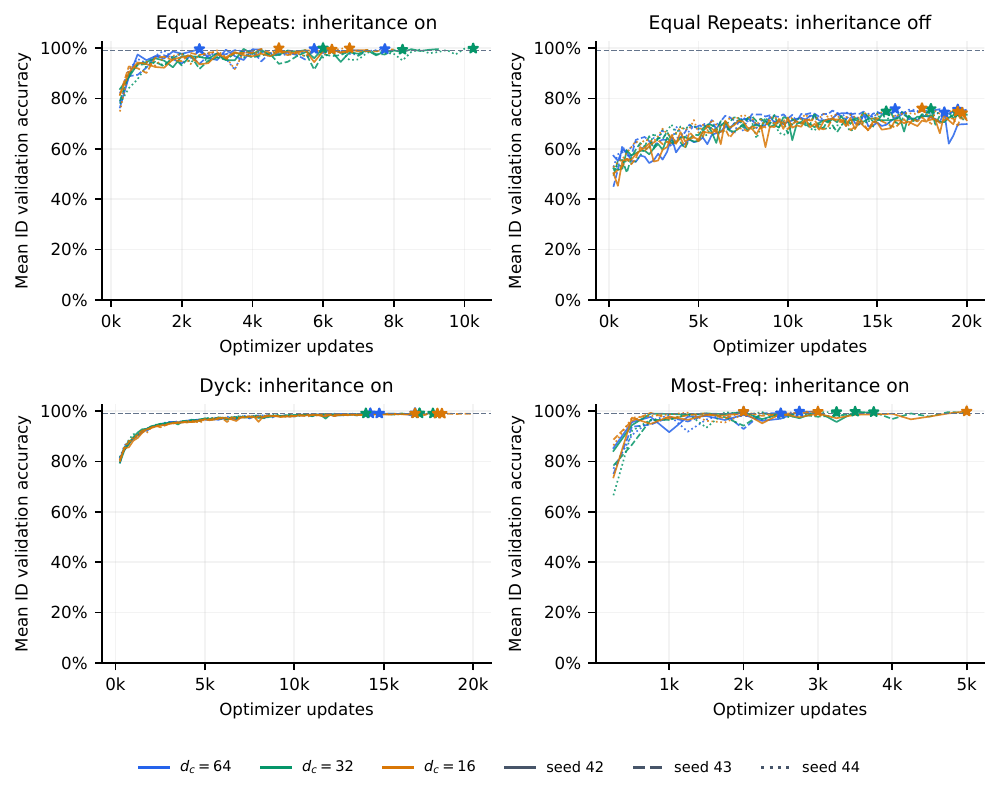}
\caption[Optimization]{\textbf{Optimization histories.} Every logged ID validation mean for all 36 runs; colors indicate width and line styles indicate seed. Stars mark the checkpoint selected by ID accuracy with a cross-entropy tie-break. The dashed horizontal line is 99\%; stopping requires two consecutive passes. Curves end at their actual stopping updates. Nine Equal Repeats runs without phase inheritance and one Dyck run exhaust the budget. Metrics and horizontal scales are task-specific, and the curves are not smoothed. Table~\ref{tab:exposure} gives steps, tokens, and wall time.}
\label{fig:optimization}
\end{figure}

The reporting unit is a training seed. Means, sample standard deviations, and paired differences should be computed from seed-level metrics without mixing evaluation lengths. Runtime can differ when accuracy-based stopping is used; report both steps and processed tokens. The figures preserve logged validation points without smoothing or extending stopped runs. Loss and validation plots describe observed optimization, not a causal advantage over architectures trained on other protocols. Available historical runtime metadata identifies an NVIDIA RTX 4090. The wall-time fields include periodic validation; they are not isolated kernel-throughput measurements.

\clearpage
\setlength{\tabcolsep}{3pt}
\begin{longtable}{llrrrrrr}
\caption{Actual training exposure for all 36 runs. Best is the selected checkpoint update; final is the stopping update. Tokens are millions of physical training tokens; Most-Freq includes separator and teacher-forced output tokens. Minutes are logged training-loop wall time, including periodic ID validation but excluding final evaluation. On/off indicates phase inheritance; $^\dagger$ denotes budget exhaustion without satisfying the two-pass threshold.}\label{tab:exposure}\\
\toprule Task / inheritance & $d_c$ & Seed & Best & Final & Tokens (M) & Min. & Params \\
\midrule\endfirsthead
\toprule Task / inheritance & $d_c$ & Seed & Best & Final & Tokens (M) & Min. & Params \\
\midrule\endhead
\bottomrule\endfoot
ER / On & 64 & 42 & 2,500 & 4,250 & 39.10 & 6.7 & 295,875 \\
ER / On & 64 & 43 & 5,750 & 7,000 & 65.10 & 10.7 & 295,875 \\
ER / On & 64 & 44 & 7,750 & 7,750 & 71.52 & 12.3 & 295,875 \\
ER / On & 32 & 42 & 6,000 & 9,250 & 85.51 & 14.6 & 268,227 \\
ER / On & 32 & 43 & 8,250 & 8,250 & 76.43 & 13.0 & 268,227 \\
ER / On & 32 & 44 & 10,250 & 10,250 & 94.80 & 15.9 & 268,227 \\
ER / On & 16 & 42 & 6,250 & 7,250 & 66.70 & 11.5 & 256,707 \\
ER / On & 16 & 43 & 4,750 & 8,000 & 74.14 & 12.4 & 256,707 \\
ER / On & 16 & 44 & 6,750 & 6,750 & 62.39 & 10.6 & 256,707 \\
ER / Off & 64 & 42 & 18,750 & 20,000$^\dagger$ & 184.83 & 31.0 & 295,875 \\
ER / Off & 64 & 43 & 16,000 & 20,000$^\dagger$ & 184.72 & 31.1 & 295,875 \\
ER / Off & 64 & 44 & 19,500 & 20,000$^\dagger$ & 183.97 & 30.8 & 295,875 \\
ER / Off & 32 & 42 & 19,500 & 20,000$^\dagger$ & 184.83 & 31.6 & 268,227 \\
ER / Off & 32 & 43 & 18,000 & 20,000$^\dagger$ & 184.72 & 30.5 & 268,227 \\
ER / Off & 32 & 44 & 15,500 & 20,000$^\dagger$ & 183.97 & 30.9 & 268,227 \\
ER / Off & 16 & 42 & 19,500 & 20,000$^\dagger$ & 184.83 & 30.7 & 256,707 \\
ER / Off & 16 & 43 & 19,750 & 20,000$^\dagger$ & 184.72 & 32.0 & 256,707 \\
ER / Off & 16 & 44 & 17,500 & 20,000$^\dagger$ & 183.97 & 31.5 & 256,707 \\
Dyck / On & 64 & 42 & 16,750 & 17,000 & 156.20 & 34.2 & 297,096 \\
Dyck / On & 64 & 43 & 14,750 & 14,750 & 136.04 & 29.7 & 297,096 \\
Dyck / On & 64 & 44 & 14,250 & 14,250 & 131.76 & 28.5 & 297,096 \\
Dyck / On & 32 & 42 & 17,000 & 17,000 & 156.20 & 33.8 & 269,448 \\
Dyck / On & 32 & 43 & 17,750 & 17,750 & 163.50 & 35.3 & 269,448 \\
Dyck / On & 32 & 44 & 14,000 & 14,250 & 131.76 & 28.8 & 269,448 \\
Dyck / On & 16 & 42 & 16,750 & 17,000 & 156.20 & 34.1 & 257,928 \\
Dyck / On & 16 & 43 & 18,250 & 20,000$^\dagger$ & 184.55 & 40.5 & 257,928 \\
Dyck / On & 16 & 44 & 18,000 & 18,000 & 165.95 & 36.8 & 257,928 \\
MF / On & 64 & 42 & 2,750 & 3,000 & 38.06 & 5.5 & 296,454 \\
MF / On & 64 & 43 & 2,500 & 2,500 & 31.71 & 4.8 & 296,454 \\
MF / On & 64 & 44 & 2,000 & 2,750 & 35.27 & 5.1 & 296,454 \\
MF / On & 32 & 42 & 3,500 & 3,750 & 47.56 & 7.0 & 268,806 \\
MF / On & 32 & 43 & 3,750 & 5,000 & 63.44 & 9.4 & 268,806 \\
MF / On & 32 & 44 & 3,250 & 3,250 & 41.63 & 5.9 & 268,806 \\
MF / On & 16 & 42 & 5,000 & 5,000 & 63.61 & 9.1 & 257,286 \\
MF / On & 16 & 43 & 2,000 & 3,000 & 37.99 & 5.5 & 257,286 \\
MF / On & 16 & 44 & 3,000 & 3,000 & 38.43 & 5.5 & 257,286 \\
\end{longtable}

\end{document}